\documentclass{article}

\usepackage{arxiv}

\usepackage[utf8]{inputenc} 
\usepackage[T1]{fontenc}    
\usepackage{hyperref}       
\usepackage{url}            
\usepackage{booktabs}       
\usepackage{amsfonts}       
\usepackage{nicefrac}       
\usepackage{microtype}      
\usepackage{lipsum}		
\usepackage{graphicx}
\usepackage{natbib}
\usepackage{doi}

\usepackage{graphicx}
\usepackage[libertine,libaltvw,liby]{newtxmath} 
\usepackage[bb = esstix, cal=zapfc,scr=rsfso]{mathalfa}
\usepackage{bm}
\usepackage{mathcommand}

\usepackage{booktabs, tabularx, multicol, multirow, caption, wrapfig, floatrow} 
\usepackage{color, colortbl, xcolor}
\usepackage[ruled,vlined]{algorithm2e}
\usepackage{enumitem}
\newcommand{\ie}{\textit{i.e.}\hspace{4pt}}

\LoopCommands\lettersUppercase[#1]
{\declaremathcommand#2{\mathcal#1}}

\LoopCommands\lettersUppercase[bb#1]
{\newmathcommand#2{\mathbb#1}}

\LoopCommands\lettersUppercase[bf#1]
{\newmathcommand#2{\mathbm#1}}

\LoopCommands\lettersUppercase[scr#1]
{\newmathcommand#2{\mathscr#1}}

\newcommand{\mathsc}[1]{\textsc{\scalefont{1.2}#1}}
\LoopCommands\lettersLowercase[sc#1]
{\newmathcommand#2{\mathsc#1}}

\usepackage{caption}
\usepackage{subcaption}

\title{Improving the Quality Control of Seismic Data through Active Learning}

\author{Mathieu Chambefort \\
	CGG - Mines ParisTech \\
	27 Avenue Carnot \\
	91300 Massy, France\\
	\texttt{mathieu.chambefort@cgg.com} \\
	\And
    Raphaël Butez \\
	Université de Lille \\
	Département Mathématiques Cité Scientifique\\
    59009 Villeneuve-d'Ascq, France \\
	\texttt{raphael.butez@univ-lille.fr} \\
	\AND
	Emilie Chautru \\
	MINES ParisTech, PSL University\\ Centre de géosciences \\
	77300 Fontainebleau, France \\
	\texttt{emilie.chautru@mines-paristech.fr} \\
	\And
	Stephan Clémençon \\
	Telecom Paris, LTCI, Institut Polytechnique de Paris \\
	19 place Marguerite Perey \\
	91120 Palaiseau, France \\
	\texttt{stephan.clemencon@telecom-paris.fr} 
}

\begin{document}
\maketitle

\begin{abstract}
In image denoising problems, the increasing density of available images makes an exhaustive visual inspection impossible and therefore automated methods based on machine-learning must be deployed for this purpose. 
This is particulary the case in seismic signal processing. Engineers/geophysicists have to deal with millions of seismic time series. Finding the sub-surface properties useful for the oil industry may take up to a year and is very costly in terms of computing/human resources. In particular, the data must go through different steps of noise attenuation. Each denoise step is then ideally followed by a quality control (QC) stage performed by means of human expertise. To learn a quality control classifier in a supervised manner, labeled training data must be available, but collecting the labels from human experts is extremely time-consuming. We therefore propose a novel active learning methodology to sequentially select the most relevant data, which are then given back to a human expert for labeling. Beyond the application in geophysics, the technique we promote in this paper, based on estimates of the local error and its uncertainty, is generic. Its performance is supported by strong empirical evidence, as illustrated by the numerical experiments presented in this article, where it is compared to alternative active learning strategies both on synthetic and real seismic datasets.

\end{abstract}



\section{Introduction}

Our goal is to perform noise detection, which are very structured and contain information on the composition of the subsurface. To this end, we present a new active learning algorithm that can perform this task after being manually trained by an expert, with minimal training time. 
The "images" on which we want to perform noise detection are the concatenation of seismic times series into a coherent image. Each time series is the measure by a sensor of the response of the subsurface to an artificial sound source at a particular geographical point. They are called "shot point gather" in geophysics  research field. Those images are then used within an inverse problem to obtain a 3D "images" of the subsurface and derived its properties (porosity, velocity...) \citep{barclay2008}.
Providing accurate properties of the subsurface, for instance below the oceans \citep{yilmaz1}, requires seismic data, what we denote images in our approach, to go through a large number of signal processing steps, so that they can be efficiently used within the inverse problem.
Among these steps, tasks dedicated to the removal of various types of noise are particularly important and subject to errors due to their inherent complexity. 
They are handled by various specific algorithms that are cascaded within workflows and all contain user-defined parameters tuned by humans (called processing geophysicists), such that the quality of the results is controlled with respect to these parameters in order to minimize the residual noise. These quality controls (QCs) are often done manually \citep{castelao2021}, which is demanding due to the large volume of the seismic data (terabytes of data). 
In this paper, the problem of automatically detecting residual noise in seismic processed images in our case with the objective of controlling the quality of the denoised images, is considered from a machine-learning perspective.
Specifically, we investigate how to apply machine-learning techniques to train a classifier detecting the presence of noise or residual noise. This is undeniably a Big Data issue in every sense of the term, insofar as billions of images are currently available for each seismic survey. A Seismic image have dimensions of at least $500\times2500$ pixels. Whereas the problem of learning a predictive rule to detect the presence of possible noise, using a preprocessed seismic image as input, can be naturally cast as a multi-class classification problem, the performance that can be achieved is conditioned by the availability of a sufficiently large number of labeled data containing both well and poorly denoised images. Obtaining a large number of labeled images is extremely difficult and costly.

The goal of this paper is to present a new method to build a classifier for noise detection that uses as little training data as possible. We present a new active learning technique, which allows us to find the most relevant training images in order to improve our classifier. By doing so, we iteratively improve the classifier by enhancing our training set with the most relevant data. As a consequence, the classifier trained on this adapted dataset has very good performance, even though it was trained on little data. Finally, this procedure will allow us to continue improving our noise detection algorithm as new data is obtained.

This paper is organized as follows: We start by first presenting the very specific structure of the seismic images and then give an overview of the machine learning modeling of the problem. In a second step, we present our active learning method and its algorithms. Finally, we test our method on several datasets: one proprietary dataset of seismic images and one classical machine learning dataset that is well suited for this kind of classification problem.

\section{Seismic data and machine learning background}

In the seismic experiments considered here, a sound source is towed by a boat and generates waves that will propagate in the subsurface and be reflected back to the surface when the waves encounter discontinuities in the subsurface. These waves are then recorded by sensors regularly distributed along cables located near the surface of the water that are also towed by a boat. Those time series then go through a complicated processing sequence \citep{mari1997} to obtain properties of the subsurface \citep{yilmaz1}.

\begin{figure*}[h!]
    \centering
    \includegraphics[width=0.3\textwidth]{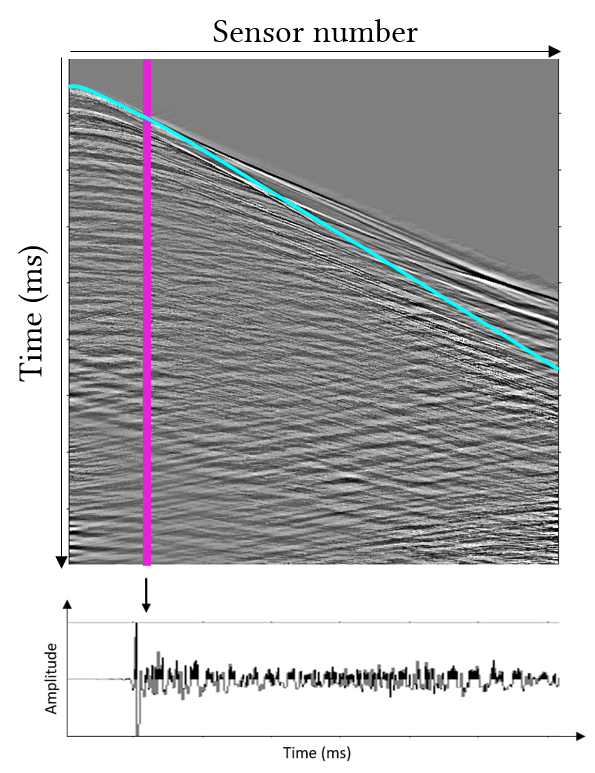}
    \includegraphics[width=0.6\textwidth, height=8cm]{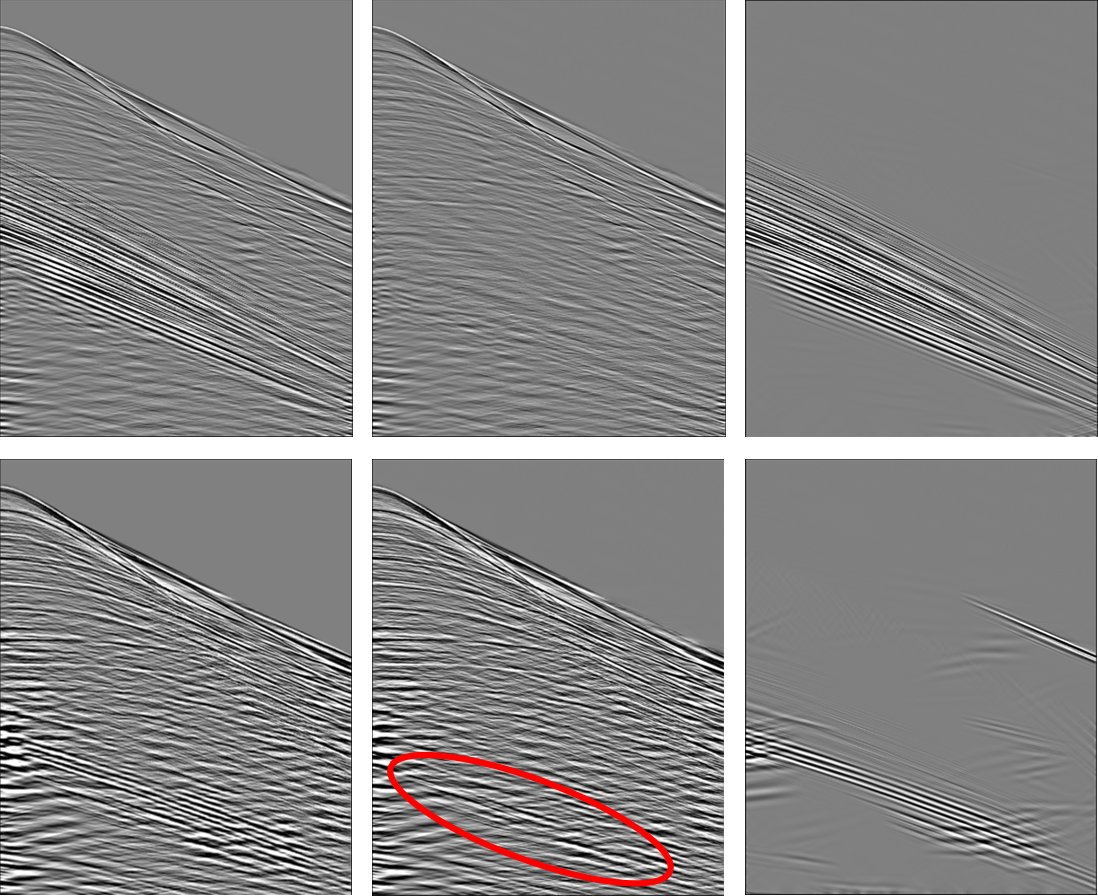}
    \caption{On the left : top, A seismic image, composed of many seismic time series, corresponding to the recordings of many sensors at one particular location. Bottom: One seismic time series, \ie, one sensor. On the right : two examples (top versus bottom) of an original image (left), its denoised version (center) and the corresponding extracted noise (right): the denoise process on top is efficient, while that on the bottom exhibits some residual noise (red circled area)}
    \label{fig:1}
\end{figure*}

In the seismic image shown in \autoref{fig:1}, each column represents the amplitude measured by a sensor over time and can be read from top to bottom (increasing recording time), with grey scale representing the amplitude of the reflected waves. Two adjacent columns correspond to two adjacent sensors along a cable measuring a very similar seismic response. This leads to a coherent and colored (with respect to the amplitude) seismic image, as displayed in \autoref{fig:1}.


The blue curve on this image represents the wave reflection of the sea floor.
These images are usually made up of $10^6$ pixels. 
Finally, a seismic survey usually covers $10,000$km$^2$ and contains millions of such images, \ie billions of time series and can easily require $40$ Tera octets of storage capacity.

\subsection{Seismic processing}\label{sec:SeismicProc}

Seismic processing consists of finding accurate properties of the subsurface from the raw seismic time series records \citep{yilmaz1, taner1979}.
However, while the recorded images contain signal (\ie, desired seismic events), they will also contain noise, which may compromise the inverse problem to find the properties of the subsurface. In practice, the seismic processing sequence is made up of tens of numerical processing steps applied to the data to prepare it for the inverse problem that will lead to certain properties of the subsurface usable for the interpretation. In the following, we will only focus on the process of noise removal.
 Before and after examples are displayed in \autoref{fig:1}. The cleaning process is heavy in human and computer time and should be performed with maximum care. 
The issue of the processing sequence is that if some noise remains in the data at a particular processing step, or if a processing step creates an issue with the data, it can affect adjacent time series during the processing flow and lead to an unrealistic representation of the subsurface. Should an issue on the data be detected a posteriori, the entire process will have to be rerun from scratch, entailing both a considerable schedule delay and extra machine and human cost.




\subsection{Machine learning and noise detection}
Following the statistical learning paradigm \citep{hastie2001}, generic seismic data, \ie shot gather image in our case, is viewed as a random pair $(X,Y)$ of input and output, with joint \textit{unknown} distribution $\bbP_{(X,Y)}$. The input $X \in \mathbb{R}^d$ corresponds to the seismic image (\ie the concatenation of times series) and the output $Y \in \{0,1\}$ to a quality label. It equals $0$ if the image passes the quality control check, \ie if there is no residual noise in $X$, and equals $1$ as soon as the presence of residual noise is suspected. In that case, the image should be evaluated by an expert.
The dataset, indexed by $\I \subset \bbN$ with cardinal $|\I|$, is modeled as a collection $(X_i,Y_i)_{i \in \mathcal{I}}$ of independent copies of the random vector $(X,Y)$. The index set $\mathcal{I}$ can be decomposed as $\mathcal{I}= \mathcal{L} \cup \mathcal{T} \cup \mathcal{U}$, where $\mathcal{L}$ and $\mathcal{T}$ respectively identify the \textit{learning set} and the \textit{test set}, both made of labeled data, while $\U$ designates the unlabeled observations.

The goal in machine learning is to find a function $g : \mathbb{R}^d \to \{0,1\}$, called a classifier, that accurately predicts $Y$ given $X$. It is generally constructed from the training dataset to be optimal in some sense. Then, its performance is evaluated through its \textit{risk}, defined as
\begin{equation*}
   \mathcal{R}_g := \mathbb{E}( \ell(Y,g(X)) )
\end{equation*}
where $\ell(y,y')\{0,1\} \times \{0,1\} \to \{0,1\} = 1_{y \neq y'}$.
In practice, since it relies on the unknown distribution $\mathbb{P}_{(X,Y)}$, it is replaced by its \textit{empirical} version, computed from the test set:
\begin{equation*}
    \widehat{\mathcal{R}}_g := \frac{1}{|\mathcal{T}|} \sum_{i \in \mathcal{T}} \ell (Y_i,g(X_i)).
\end{equation*}
In order to improve the classifier, the training set can be extended by labeling some elements of $\U$ and adding them to $\mathcal{L}$. They are usually picked at random, uniformly in $\mathcal{U}$. Proceeding so guarantees convergence towards an optimal classifier as the number of labeled data tends to infinity. Nevertheless, it may take a lot of labeled data to reach an efficient, satisfactory classifier.


In the present situation, labeling seismic data is very costly. Thus, improving the speed of convergence of the classifier towards its equilibrium is paramount: the amount of labeled data needed to get a sufficient result should be as small as possible. To acheve this, the \textit{active learning} paradigm proposes selecting the most relevant data in the unlabeled set, instead of picking these new observations purely at random.

\section{Our method:  local risk-based active learning strategy}\label{sec:local_al}
We present our solution to the noise detection problem, which includes efficiently building a classifier thanks to a new active learning strategy.

\subsection{Optimal classifier: minimizing the estimated local risk}
In order to implement an active learning scheme, the training set needs to be enhanced with the data that has maximum influence on risk minimization. We adopt an approach based on the \textit{local risk function}, defined for a classifier $g$ and a loss function $\ell$ as the mapping:
\begin{equation*}
    x \in \bbR^d \mapsto R_{g}(x) := \mathbb{E}(   \ell(Y,g(X) )\mid X=x ).
\end{equation*}
It is related to the risk $\mathcal{R}_{g}$ through the chain rule
\begin{equation*}
    \mathcal{R}_{g} = \mathbb{E}(\ell(Y,g(X))) = \mathbb{E}(\mathbb{E}(\ell(Y,g(X))\mid X)) = \mathbb{E}( R_{g}(X) ). 
\end{equation*}

To this end, the unlabeled data that maximizes the local risk function would be selected and experts would be asked to label it. Then, the corresponding input-output couple would be added to the training set, with index
\[ 
    i_{\ast} := \underset{{i \in \mathcal{U}}}{\mathrm{argmax}}\ R_{g}(X_i).    
\]

Intuitively, thinking of the graph of the local risk function as a surface in $\mathbb{R}^d \times [0,1]$, this amounts to selecting its high peaks and setting them to $0$ in the next update of the classifier. This approach is expected to work under the condition that, when adding a point to the training set, the local risk function does not change drastically, and new peaks do not appear (thereby avoiding a \textit{whack-a-mole} configuration). This should be the case for \textit{local} classifiers such as $k$-nearest-neighbors or random forests. This approach is not expected to perform well with non-local classifiers, such as logistic regression.
Choosing a point according to the $L^{\infty}$ norm to minimize the $L^1$ norm may be surprising at first glance. The distribution $\mathbb{P}_X$ is unknown. If one approximates $\bbP_X$ by the empirical distribution of the $X_i$'s for $i \in \U$ 
$  \widehat{\bbP_X} = \frac{1}{|\U|} \sum_{i \in \U} \delta_{X_i},$
then the risk of a classifier $g$ can be written as 
\[ \mathcal{R}_g \approx \frac{1}{|\U|} \sum_{i \in \U} R_h(x_i).  \]
With this representation, the most relevant data to label is the one maximizing the local risk.

Unfortunately, this approach cannot be performed exactly, as it relies on the unknown distribution $\mathbb{P}_{(X,Y)}$. The vector $(R_g(X_i))_{i\in\U}$ of the local risk function taken at each unlabeled data point is replaced by an estimator $(\widehat{R_g(X_i)})_{i\in\U}$. Incidentally, $i_\ast$ is redefined as
\begin{equation}\label{eq:inew}
    i_{\ast} := \underset{{i \in \mathcal{U}}}{\mathrm{argmax}}        \text{ }\widehat{R_{g}(X_i)}.    
\end{equation} 
When several indices reach this maximum, $i_\ast$ is picked at random, uniformly among them.
A plethora of estimators could be chosen to assess the local risk function. We opted for the Nadaraya-Watson (also called local-constant) estimator, for both its non-parametric and local qualities; using a local estimator for a local classifier seems sensible. Given a window parameter $h \in \bbR^+$ and a kernel function $K_h$ (typically the standard Gaussian kernel $(x,x^\prime) \in \bbR^d\times\bbR^d \mapsto e^{-\|x-x'\|^2/h}$), this estimator of the local risk function at point $x \in \bbR^d$ is defined as
\begin{equation*}
    \widehat{R_{g}(x)} := \sum_{i \in \L} \ell(Y_i,g(X_i))\,\dfrac{K_h(x,X_i)}{\sum\limits_{j\in\L} K_h(x,X_j)}.
\end{equation*}
In the sequel, when necessary, its dependence on $h$ and $\L$ shall be made explicit by writing $\widehat{R_{g}}(x; h, \L)$.

\subsection{Batch approach and survey schemes}

The seismic data experts who will label new observations, are humans (geophysicists). From a practical point of view, it is sub-optimal to occasionally solicit them to label a single image. Taking this into account, our procedure is readily adapted to the case where batches of $K \in \bbN$ images are added to the training set at each update of the classifier. After consulting with the geophysics experts we work with, it was decided to set $K \approx 20$ (this task is completed in about $15$ minutes). 

Alternately, the selection process could be made random to favor the exploration of the input space. In the vein of \cite{CBCP19} or \cite{beygelzimer2009}, a conditional Poisson sampling design could be used to that effect. The general idea is to give priority to the unlabeled data with high estimated local risk, but still allow the others to be picked once in a while. Precisely, each candidate $i \in \U$ is given a sampling weight $\pi_i \in [0,1]$, named \textit{inclusion probability}, such that $\sum_{i\in\U} \pi_i = K$. Then, it is selected independently according to the result of a Bernoulli trial with probability $\pi_i$. The procedure is repeated until exactly $K$ elements of $\U$ are selected (\autoref{alg:survey}). The inclusion probabilities are taken proportional to the estimated local risk (possibly augmented by its variance):
\begin{equation}\label{eq:incpr}
    \forall\,i\in\U, \quad \pi_i := K\,\dfrac{\widehat{R_g(X_i)}}{\sum\limits_{j\in\U}\widehat{R_g(X_j)}}.
\end{equation}
It can happen that some number $K_0$ of them exceed $1$. In that case, the corresponding data are automatically chosen and the process is restarted with the remaining data, using $K-K_0$ instead of $K$ in \eqref{eq:incpr}.


\section{Algorithms}
In this section, we give all the algorithms to precisely implement our methods, in the same order as presented above.
We start with the empirical risk minimization, where the parameter $h$ in the kernel is computed thanks to the classical \textit{leave-one-out cross-validation} method.

\begin{procedure}[]{}
\SetAlgoLined
\SetKwInOut{KwIn}{Input}
\SetKwInOut{KwOut}{Output}
\KwIn{$(X_i,Y_i)_{i \in \L}$ , a classifier $g$}
\KwOut{$h \in \H$, where $\H$ is a fixed grid in $\mathbb{R^+}$}
\For{$h \in \H$ } {
$E(h) \gets \frac{1}{|\L|} \sum\limits_{j \in \L} \Big( \ell(Y_j,g(X_j)) -  \widehat{R_g}(X_j;h,\L \setminus \{j\})  \Big)^2 $}
\Return{$h \gets  \underset{{h \in \H}}{\mathrm{argmin}}\ E(h)$}
\caption{Loocv()}
\end{procedure}
We are now ready to give the main algorithms for our active learning scheme.
\begin{algorithm}[h!]{}
\SetAlgoLined
\SetKwInOut{KwIn}{Input}
\SetKwInOut{KwOut}{Output}
\KwIn{$(X_i,Y_i)_{i \in \L \cup \U}$, a classifier $g$, $K \in \mathbb{N}$, $\lambda \in \mathbb{R}^+$}
\KwOut{$i_\ast^{(1)}, \dots, i_\ast^{(K)} \in \U$}
$h \gets\Loocv((X_i,Y_i)_{i \in \L},g)$\\
\For{$i \in \U$}{
$R_i \gets {\widehat{R_{g}}(X_i;h,\L)} + \lambda \widehat{V_g}(X_i, (X_j,Y_j)_{j \in \L}, g) $ }
$\U_0 \gets \U$

\For{$p \in \{1, \dots, K\}$}{
$i_\ast^{(p)} \gets \texttt{chosen uniformly in } \mathrm{argmax}_{i \in \U_{p-1} } R_i$  \\
$\U_p \gets \U_{p-1} \setminus \{ i_\ast^{(p)}\}$
}
\Return{  $ i_\ast^{(1)}, \dots, i_\ast^{(K)} $ }
\caption{Local risk with $K$ batches}
\end{algorithm}
\begin{algorithm}[h!]{Test}
\SetAlgoLined
\DontPrintSemicolon
\SetKwInOut{KwIn}{Input}
\SetKwInOut{KwOut}{Output}
\KwIn{$(X_i,Y_i)_{i \in \L \cup \U}$, a classifier $g$, $K \in \mathbb{N}$}
\KwOut{$I=\{i_\ast^{(1)}, \dots, i_\ast^{(K')} \} \in \U$}
$h \gets \texttt{LOOCV}((X_i,Y_i)_{i \in \L},g))$\;
$U \gets \U$\;
$R \gets $ array of the size of $U$\;
$K_0 \gets K$\;
\For{$i \in U$}{
$R[i] \gets {\widehat{R_{g}}(X_i,h,\L)}$}
$r \gets \sum_{i\in U} R[i]$\;
\If{$r=0$}{$I \gets$ $K_0$ indices, chosen uniformly in $U$}
\While{ $\exists i_0 \text{ such that } K_0 R[i_0]/r >1$}{
    \For{$i \in U$}{
        \If{$K_0R[i]/r >1$}{$I \gets I \cup \{i\} $
            $U \gets U\setminus\{i\}$\;
            $K_0\gets K_0-1$}\;
            $R \gets $ array of the size of $U$\;
    }
    \For{$i \in U$}{
        $R[i] \gets {\widehat{R_{g}}(X_i,h,\L)}$
    }
    $r \gets \sum_{i\in U} R[i]$\;
    \If{$r=0$}{$I \gets$ $K_0$ indices, chosen uniformly in $U$}
}
\For{$i \in U$}{
    $\theta = \textbf{Bernoulli}(K_0R[i]/r)$  \;
    \If{$\theta=1$}{ $I \gets I \cup \{i\}$  }
}
\Return{I}
\caption{Survey Scheme with $K$-batches}\label{alg:survey}
\end{algorithm}

\section{Experiments}
In this section, we test our algorithms on two datasets: a seismic dataset, which is the property of CGG and is not available publicly, but is our main interest, and the Wisconsin Breast Cancer (WBC) dataset \citep{dua2017}, available on the UCI website, and which is widely used in machine learning. We chose the WBC dataset because it is well suited for binary classification, with high dimensional data and a large amount of data. With low dimensional data, all the classical algorithms converge very quickly towards the equilibrium, and there is not much room for improvement of these methods.

All the algorithms described above are expected to perform well if one starts from a non-empty training set. If one starts with no labeled data, then the local error is poorly estimated, and our method is not better than the random selection. Starting with a small amount of manually labeled data seems to be a reasonable requirement for applications, where it is common to manually label some data before considering the construction of a classifier.
This is why we started the experiments with $300$ labeled data for the seismic dataset and $50$ labeled data for the WBC dataset.
In addition, we focus on the case where the data are labeled by batches of $20$ images. This batch situation is the most natural for industrial applications and is the one we will focus on.

\begin{figure*}[h!]
    \centering
    \begin{subfigure}[b]{\textwidth}
    \includegraphics[width=0.48\textwidth, height=5cm]{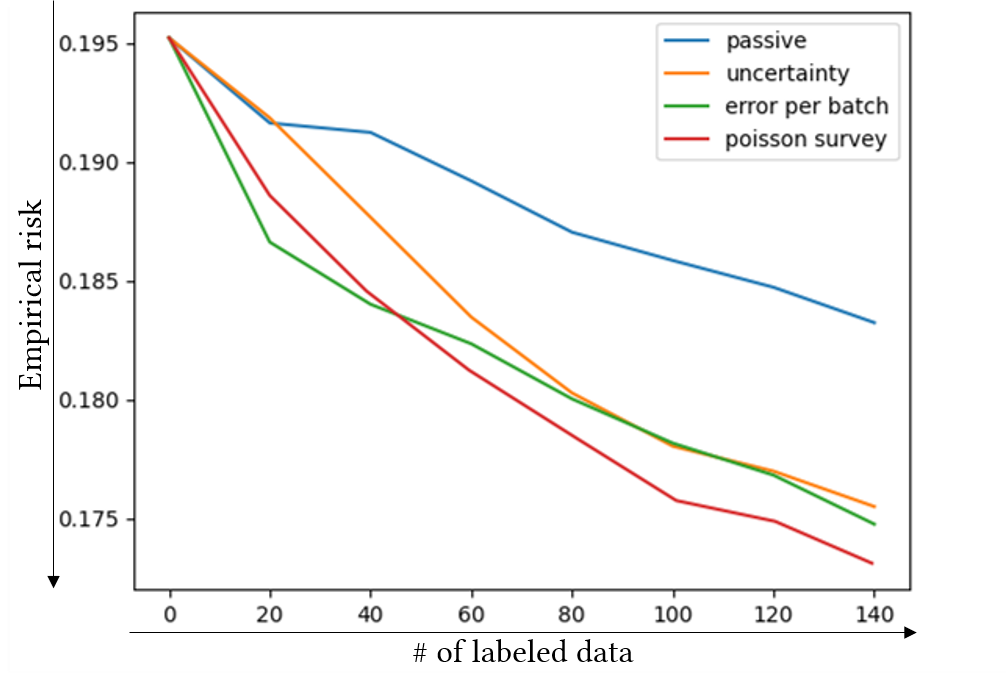}
    \includegraphics[width=0.48\textwidth, height=5cm]{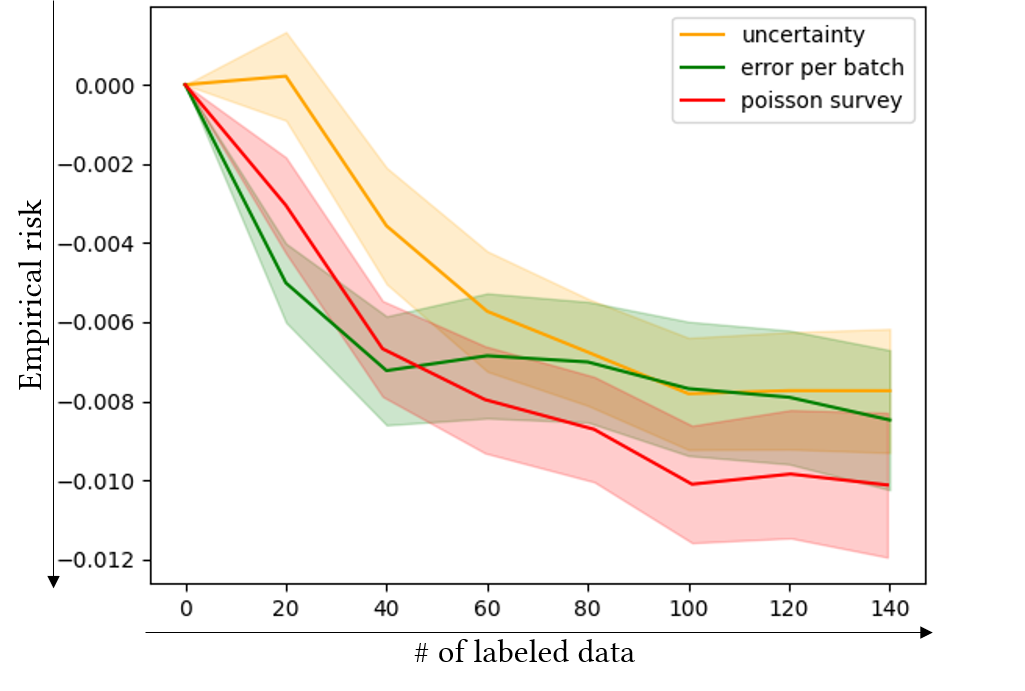}
    \caption{Seismic dataset with a warm-up of $|\L|=304.$}
    \label{fig:warm_up_seismic}
    \hfill
    \end{subfigure}
    \begin{subfigure}[b]{\textwidth}
    \includegraphics[width=0.48\textwidth, height=5cm]{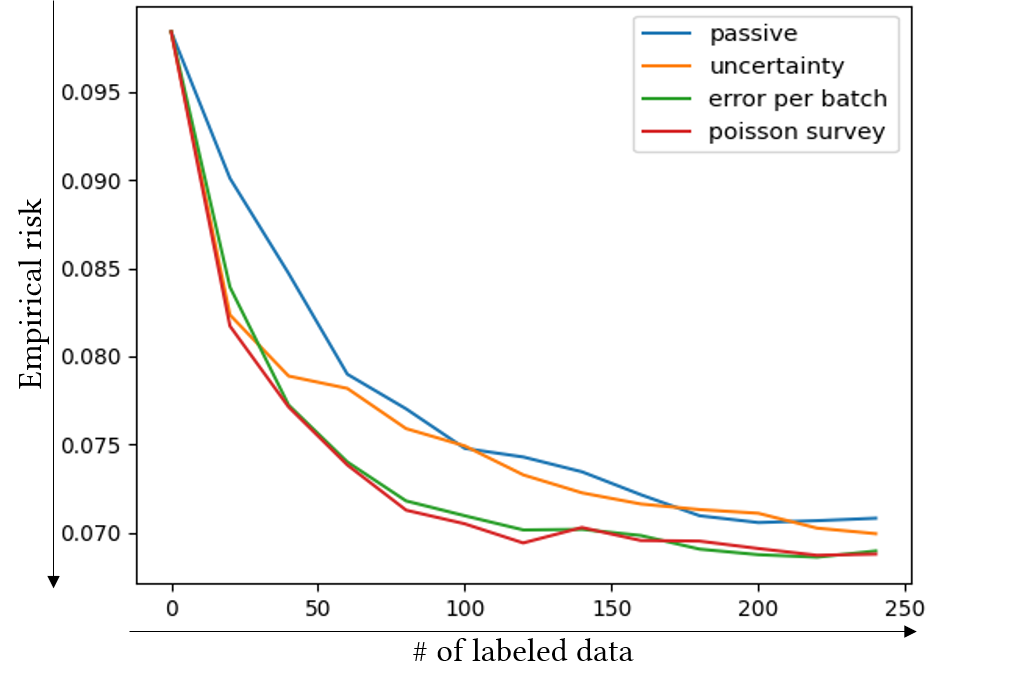}
    \includegraphics[width=0.48\textwidth, height=5cm]{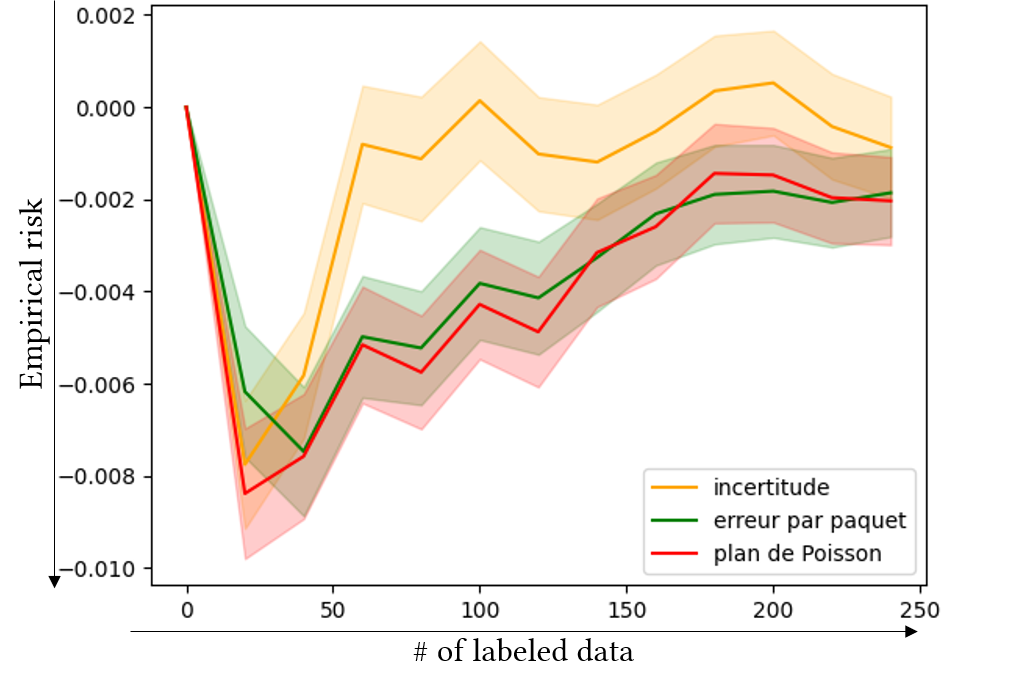}
    \caption{Wisconsin Breast Cancer dataset with a warm-up of $|\L|=50$.}
    \label{fig:warm_up_wbc}
    \end{subfigure}
    \hfill

    \caption{
    Our approaches using local risk per batch (in green) and the Poisson plan survey (in red) compared to uncertainty sampling (orange) and passive learning (blue).
    Each curve on the left is obtained by averaging 500 simulations. On the right we can observe the average and variance of each tested method with the passive learning approach.}
  \label{fig:warmup_seismic}
\end{figure*}

In Figure \ref{fig:warm_up_seismic}, we present the performance of several active learning heuristics with the k-NN classifier \citep{hastie2001} and applied to the seismic data presented earlier. The left image shows the average risk of the classifiers built with different learning heuristics, each of the experiments being repeated $500$ times. In this figure, several algorithms are considered: the blue curve corresponds to passive learning, where the points are chosen uniformly at random, the orange curve corresponds to the uncertainty algorithm \cite{tong2002, yang2018}, which is known to be the best active learning strategy in many cases of low dimension, the green curve corresponds to our local error strategy by batches and the red curve corresponds to our local error strategy couple with a Poisson plan. The image on the right in Figure \ref{fig:warm_up_seismic} shows the difference between the blue curve and the other curves. The shades of color represent the standard error of the mean (SEM) around the curve. We observe that our strategies are significantly better than the passive strategy. Even if our strategies seem slightly better than the uncertainty method, the difference falls in the range of the SEM, so our methods are not significantly better than the uncertainty method on the seismic dataset.

For this seismic dataset, it was not possible to push the simulations to the point where the curves join together, due to a very long computation time for each training.

In Figure \ref{fig:warm_up_wbc}, we present the performance of the same algorithms as in Figure \ref{fig:warm_up_seismic}, but used on the WBC dataset. The main interest of testing our methods on the WBC dataset is the reproducibility for the reader, with all the codes available at \href{ https://github.com/mathchambef/local-active-learning}{author github}.
We observe that our methods give the best performance and are significantly better than both the passive strategy and the uncertainty method.




\begin{figure}[h!]
    \centering
    \includegraphics[width=0.5\textwidth, height=5cm]{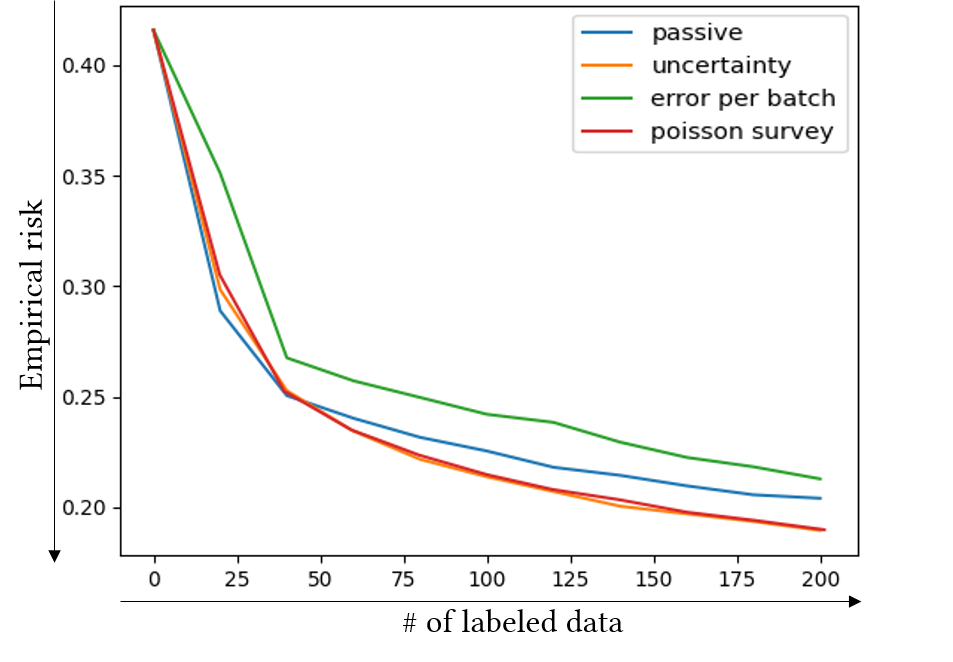}
    \caption{Seismic dataset without warmup, batches of $20$ images.}
    \label{fig:no_warm_up_seismic_b20}       
\end{figure}

In Figure \ref{fig:no_warm_up_seismic_b20}, we test our algorithm on the seismic dataset, starting from an empty labeled set. As expected, we observe that there is first a warm-up phase during which our methods are comparable to the passive learning. Because of this warm-up phase, it is hard to see if our methods are actually better than the passive learning, and the active strategies start to perform well when the whole system is already close to equilibrium. This warm-up phenomenon is common in active learning and we can also observe it for the uncertainty method.

\begin{figure}[H]
    \centering
    \includegraphics[width=0.5\textwidth, height=5cm]{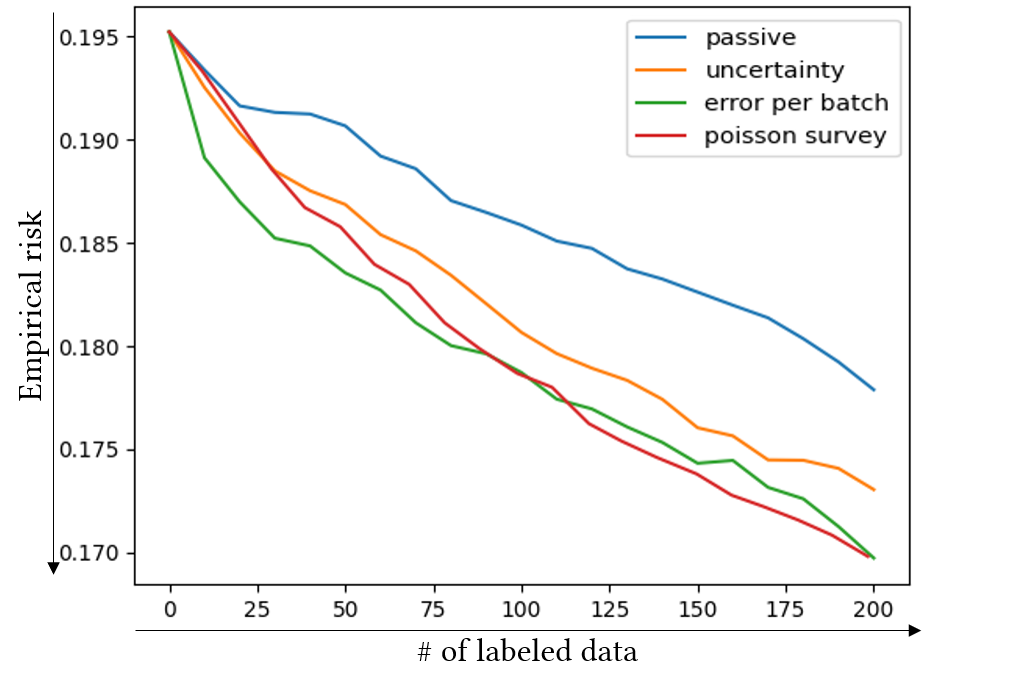}
    \caption{Seismic dataset, batches of $10$ images.}
    \label{fig:warm_up_seismic_b10}       
\end{figure}

\begin{figure}[h!]
    \centering
    \includegraphics[width=0.5\textwidth, height=5cm]{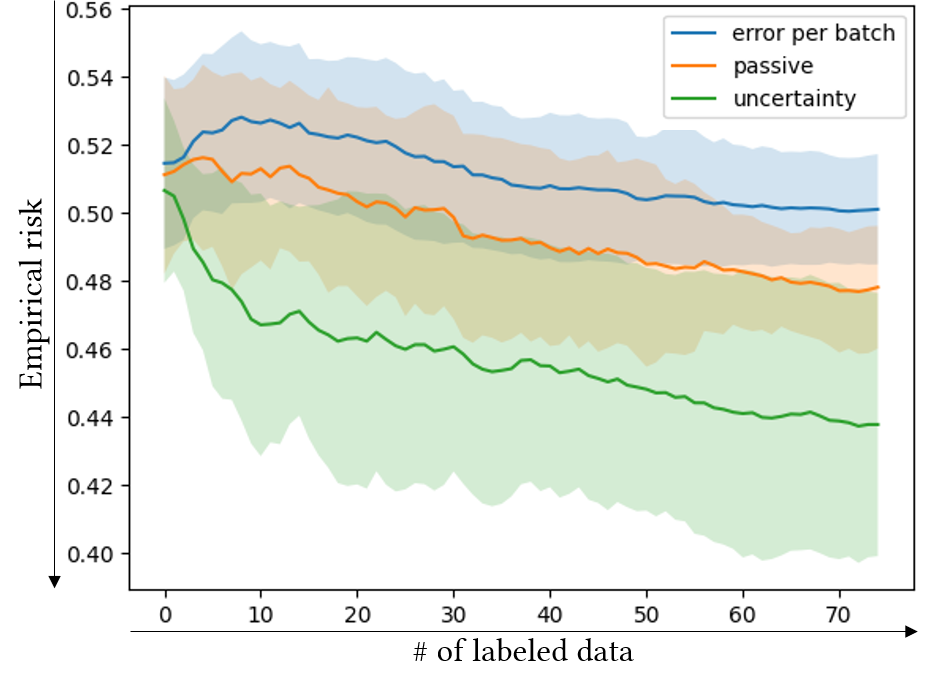}
    \caption{whack-a-mole situation}
    \label{fig:whack-a-mole}       
\end{figure}

In Figure \ref{fig:warm_up_seismic_b10} we tested the influence of the size of the batch on the performance of the algorithms and see that reducing the size of the batch does not drastically change the performance of the algorithms. For a batch of size $K=1$, the Poisson plan and the error-by-batch method should coincide exactly. We believe that for very large batches, the Poisson method should have a better exploration-exploitation compromise than the error-by-batch and should have slightly better performance.
From the beginning, we specified that our methods are designed to work with local classifiers, such as the $k$-NN or the random forest algorithm. What happens if we try our algorithms on the logistic regression, which is the opposite of a local algorithm: labeling one image changes separating hyperplane and may change the outcome of the classifier to be far away from the labeled data. In Figure \ref{fig:whack-a-mole}, we show the performance of our algorithms on the seismic dataset with the logistic regression. We are not surprised to see that our technique works poorly in this situation, while the uncertainty heuristic still has great performance. The shaded colors represent the SEM around the curves.

\section{Discussion and conclusion}

We provided a novel and original way of selecting new data to label in an active learning pool-based setting \citep{settles2010}, based on the estimation of the local risk of a classifier.
We recommend using our method  with a local classifier such as the k-NN or the random forest \citep{hastie2001} as, per construction, our method can perform poorly with non-local classifiers like the logistic regression as locally decreasing the error in a subspace of $\mathcal{X}$ can increase it somewhere else. 

A limit of the provided method is that if the batch-size $K$ is small, our survey approach will not lead to an improvement compared to the classical selection per batch based on the local error criterion. Indeed, when the inclusion probability is $1$, the points are selected automatically, which boils down to the classical batch approach. This issue can be solved with a good choice of the initial labeled training set, \ie, a good choice of the warm-up data. If the starting training set is "good" in a sense, we may start with a good estimation of the local error, and then the survey approach should be better than the local error with batches.

The classical approach consists of selecting the $K$ data that maximize the estimation of the local error and can be seen as a greedy algorithm in the bandit research area \citep{lattimore2020}. Another simple way of adding an exploration criterion would be to do a 
$\varepsilon$-greedy algorithm, where the exploitation criterion is done following our method and the exploration at random, or using some well-known exploration criterion of active learning like the mismatch first traversal \citep{zhao2020}.

Another classical exploration criterion that we can consider here could be an uncertainty-based criterion. The points that will be selected will be the ones that maximize the linear combination of our exploitation criterion and this uncertainty criterion. We have tried to add the variance of our estimator of the local error computed using a leave-one-out approach for this criterion, but in the case of our real data (that belong to a huge space), it is very costly in term of computing time.
Our novel active learning criterion is promising both for exploration and exploitation and already produces interesting results on different datasets. Our goal is to exposed this new approach and it would benefit from being compared with other classical methods \citep{yang2018} on more machine learning datasets \citep{dua2017}.

The method presented in this paper is original compared to other approaches from active learning literature: firstly, it uses the local risk as exploitation criterion, as reducing the local risk should reduce the global one if the classifier is local, and secondly, the exploration criterion method is based on survey methods.

\section{Acknowledgements}
The authors would like to thank CGG for permission to show the images. R.B. is supported by the Labex CEMPI (ANR-11-LABX-0007-01).

\section{Declaration of interests}
The authors declare that they have no known competing financial interests or personal relationships that could have appeared to influence the work reported in this paper.

This papier is under consideration at Pattern Recognition Letter.

%
%


\bibliographystyle{unsrtnat}
\bibliography{bibli}

\end{document}